\documentclass[11pt,fleqn,fancyheadings,a4paper]{article}

\usepackage{hyperref}       
\usepackage{url}            
\usepackage{booktabs}       
\usepackage{amsfonts}       
\usepackage{nicefrac}       

\usepackage{algpseudocode}

\newcommand*\Let[2]{\State #1 $\gets$ #2}

\usepackage{authblk}
\usepackage[twoside,bindingoffset=2.1cm,includeheadfoot,textwidth=17cm,textheight=26.5cm,top=0.5cm]{geometry}

\usepackage{graphicx,algorithmicx,algorithm,algpseudocode,xspace,xcolor}
\usepackage{subfigure}

\usepackage{tikz}
\usetikzlibrary{arrows,shapes,backgrounds,through,shadows,snakes,patterns,mindmap}
\usetikzlibrary{calc}
\graphicspath{{./matlab/}{./}{./figures/}{./../figures/}{./../matlab/}{./../matlab/figures/}{./../codeOO/misc/}}

\usepackage{amsmath,amssymb}

\reversemarginpar

\newcommand{\finaldelete}[1]{}

\newcommand{\bec}{\begin{center}}
\newcommand{\enc}{\end{center}}
\newcommand{\bef}{\begin{figure}}
\newcommand{\enf}{\end{figure}}

\newcommand{\ocm}{\hspace{1cm}}
\newcommand{\hcm}{\hspace{0.25cm}}

\newcommand{\marginlabel}[1]{}

\newcommand{\tcut}[1]{}
\newcommand{\jdelete}[1]{}

\newcommand{\nts}[1]{}

\newcommand{\dbeginexercise}{\begin{exercise}[\red{PLEASE DELETE THIS EXERCISE}]}

\newcommand{\red}[1]{{\color{red}#1}}

\newcommand{\dcup}[1]{}
\newcommand{\ddcup}[1]{}

\newcommand{\rcup}[1]{}

\newcommand{\ncup}[1]{}

\newcommand{\tw}{\textwidth}
\newcommand{\newl}{\hspace{1mm}\\}

\newcommand{\bmp}[1]{\begin{minipage}{#1}}
\newcommand{\bmpp}[2]{\begin{minipage}[#1]{#2}}
\newcommand{\emp}{\end{minipage}}
\newcommand{\be}{\begin{equation}}
\newcommand{\ee}{\end{equation}}
\newcommand{\bea}{\begin{eqnarray}}
\newcommand{\eea}{\end{eqnarray}}

\newcommand{\beqn}{\begin{equation}}
\newcommand{\eeqn}{\end{equation}}

\newcommand{\bueq}{\begin{equation}}
\newcommand{\eueq}{\end{equation}}

\newcommand{\cut}[1]{}

\newcommand{\br}[1]{\left( {#1} \right)}

\renewcommand{\eqref}[1]{(\ref{#1})}

\renewcommand{\eqref}[1]{equation(\ref{#1})}

\newcommand{\figref}[1]{fig(\ref{#1})}

\newcommand{\figrefm}[2]{fig(\ref{#1}{#2})}
\newcommand{\secref}[1]{section(\ref{#1})}

\renewcommand{\algref}[1]{algorithm(\ref{#1})}

\newcommand{\pdiff}[2]{\frac{\partial{#1}}{\partial{#2}}}

\renewenvironment{remark}[1][Remark]{\begin{trivlist}
\item[\hskip \labelsep {\bfseries #1}]}{\end{trivlist}}

\newcommand{\trans}{^{\textsf{T}}}

\newcommand{\aside}[1]{}

\renewcommand{\eqref}[1]{equation (\ref{#1})}

\newcommand{\appref}[1]{Appendix (\ref{#1})}

\newcommand{\bvb}{\begin{alltt}}
\newcommand{\evb}{\end{alltt}}
\newcommand{\beq}{\begin{equation}}
\newcommand{\eeq}{\end{equation}}

\definecolor{lightgreen}{rgb}{0,    1.0000,    0}
\definecolor{newlightgreen}{rgb}{0.86,0.86,0.99}

\definecolor{lightred}{rgb}{1,0.92,0.92}
\definecolor{verylightred}{rgb}{1.0,0.85,0.85}
\definecolor{truelightgreen}{rgb}{0.9,0.9,0.9}
\definecolor{purple}{rgb}{0.5,0.5,1}
\definecolor{darkgreen}{rgb}{0.3,1,0.3}

\tikzstyle{cont}=[circle,draw=blue!50,thick,minimum size=6mm,line width=2pt,>=stealth]  
\tikzstyle{neur}=[rectangle,draw=green!50,fill=green!50,minimum size=6mm,line width=2pt,>=stealth]  
\tikzstyle{chapter}=[rectangle,draw=green!50,fill=green!50,minimum size=6mm,line width=2pt,>=stealth,draw=none,fill=none,right]  
\tikzstyle{ocont}=[ellipse,draw=blue!50,thick,minimum size=6mm,line width=2pt,>=stealth]  
\tikzstyle{blackcont}=[circle,draw=black!50,thick,minimum size=6mm,line width=2pt,>=stealth]  
\tikzstyle{oval}=[ellipse,draw=blue!50,thick,minimum size=6mm,line width=2pt,>=stealth]  
\tikzstyle{disc}=[rectangle,draw=blue!50,thick,minimum size=6mm]  
\tikzstyle{obs}=[fill=blue!20,thick]  
\tikzstyle{marg}=[fill=north east lines,thick]

\tikzstyle{gm}=[fill=blue!20,thick,draw=blue!50,circle,minimum size=4mm]  
\tikzstyle{ngm}=[fill=none,thick,draw=blue!50,circle,minimum size=4mm]  
\tikzstyle{ml}=[fill=blue!20,thick,draw=blue!50,circle,minimum size=4mm]  
\tikzstyle{nml}=[fill=none,thick,draw=blue!50,circle,minimum size=4mm]  
\tikzstyle{ai}=[fill=blue!20,thick,draw=blue!50,circle,minimum size=4mm]  
\tikzstyle{nai}=[fill=none,thick,draw=blue!50,circle,minimum size=4mm]  
\tikzstyle{ts}=[fill=blue!20,thick,draw=blue!50,circle,minimum size=4mm]  
\tikzstyle{nts}=[fill=none,thick,draw=blue!50,circle,minimum size=4mm]  
\tikzstyle{pm}=[fill=blue!20,thick,draw=blue!50,circle,minimum size=4mm]  
\tikzstyle{npm}=[fill=none,thick,draw=blue!50,circle,minimum size=4mm]  

\tikzstyle{celim}=[circle,draw=red!25,thick,minimum size=6mm,line width=2pt,>=stealth]  %
\tikzstyle{delim}=[draw=red!25]  %

\tikzstyle{fillred}=[fill=red!20,thick]  
\tikzstyle{fillgreen}=[fill=green!20,thick]  
\tikzstyle{purered}=[fill=red]  
\tikzstyle{state}=[rectangle,fill=red!20]  
\tikzstyle{sobs}=[fill=green!15,thick]  
\tikzstyle{fact}=[fill,minimum size=1.5mm,line width=2pt,>=stealth]
\tikzstyle{varfact}=[draw,minimum size=1.5mm,line width=2pt,>=stealth]
\tikzstyle{sep}=[rectangle,draw=magenta!50,thick,minimum size=6mm]  
\tikzstyle{det}=[fill=red!15,rectangle,draw=red!50,thick,minimum size=6mm]  

\tikzstyle{dethid}=[diamond,draw=red!50,thick,minimum size=6mm]  

\tikzstyle{lineball}=[fill,-*,draw=red!50,line width=1.5pt]
\tikzstyle{redball}=[mark=*,mark options={fill=red!50,draw=red},mark size=0.5pt]
\tikzstyle{greenball}=[mark=*,mark options={fill=green!50,draw=green},mark size=0.5pt]
\tikzstyle{hid}=[circle,draw,thick]  

\tikzstyle{dec}=[rectangle,draw=red!50,thick,minimum size=6mm]  
\tikzstyle{opt}=[star,draw=red!50,thick,minimum size=6mm]  
\tikzstyle{utility}=[diamond,draw=red!50,thick,minimum size=6mm]  
\tikzstyle{contdec}=[circle,draw=blue!50,thick,fill=blue!10,line width=2pt]  
\tikzstyle{decutility}=[diamond,draw=red!50,thick,minimum size=6mm]  

\tikzstyle{contobs}+=[cont]
\tikzstyle{contobs}+=[obs]
\tikzstyle{discobs}+=[disc]
\tikzstyle{discobs}+=[obs]

\tikzstyle{obsred}+=[obs]
\tikzstyle{obsred}+=[red]

\tikzstyle{background grid}=[draw, black!50,step=.1cm]
\tikzstyle{dgraph}=[->, line width=1.5pt]
\tikzstyle{ugraph}=[line width=1.5pt]

\tikzstyle{background rectangle}=[draw=blue!50,rounded corners]

\newcommand{\btz}{\begin{tikzpicture}}
\newcommand{\etz}{\end{tikzpicture}}

\newcommand{\bigo}[1]{O\br{#1}}

\newcommand{\bi}{{\begin{itemize}}}
\newcommand{\ei}{{\end{itemize}}}

\usepackage{graphicx,algorithmicx,algorithm,algpseudocode,xspace,xcolor}
\usepackage{subfigure}

\usepackage{tikz}
\usetikzlibrary{arrows,shapes,backgrounds,through,shadows,snakes,patterns,mindmap}
\usetikzlibrary{calc}
\graphicspath{{./matlab/}{./}{./figures/}{./../figures/}{./../matlab/}{./../matlab/figures/}{./../codeOO/misc/}}

\usepackage{amsmath,amssymb}

\newcommand{\gradJ}[1]{J'\br{#1}}

\newcommand{\gradgradJ}[1]{J''\br{#1}}

\title{Nesterov's Accelerated Gradient and Momentum as approximations to Regularised Update Descent}

\author{Aleksandar Botev, Guy Lever and David Barber}

\affil{Department of Computer Science\\
University College London}
\date{\today}

\begin{document}

\maketitle

\begin{abstract}
We present a unifying framework for adapting the update direction in gradient-based iterative optimization methods. As natural special cases we re-derive classical momentum and Nesterov's accelerated gradient method, lending a new intuitive interpretation to the latter algorithm. We show that a new algorithm, which we term Regularised Gradient Descent, can converge more quickly than either Nesterov's algorithm or the classical momentum algorithm. 
\end{abstract}

\section{Introduction}

We present a framework for optimisation by directly setting the parameter update to optimise the objective function. Under natural approximations, two special cases of this framework recover Nesterov's Accelerated Gradient (NAG) descent\cite{nag} and the classical momentum method (MOM)\cite{momentum}. This is particularly interesting in the case of NAG since, though popular and theoretically principled, it has largely defied intuitive interpretation. We show that (at least for the special quadratic objective case) our algorithm can converge more quickly than either NAG or MOM.\newl

Given a continuous objective $J(\theta)$ we consider iterative algorithms to minimise $J$. We write $\gradJ{\theta}$ for the gradient of the function evaluated at $\theta$, and similarly $\gradgradJ{\theta}$ for the second derivative\footnote{These definitions extend in an obvious way to the gradient vector and Hessian in the vector $\theta$ case.}.  Our focus is on first-order methods, namely those that form the parameter update on the basis of only first order gradient information. 

\subsection{Gradient Descent}

Perhaps the simplest optimisation approach is Gradient Descent (GD) which, starting from the current parameters, locally modifies the parameter $\theta_t$ at iteration $t$ to reduce $J$. Based on the Taylor series expansion
\beq
J(\theta_t+v_t)=J(\theta_t)+v_t\gradJ{\theta_t}+\bigo{v_t^2}
\eeq
for a small learning rate $\alpha_t>0$, setting $v_t=-\alpha_t\gradJ{\theta_t}$   reduces $J$. This motivates the GD update $\theta_{t+1}=\theta_t+v_t$. 
For convex Lipshitz $J$  GD converges to the optimum value $J^*$ as $J(\theta_t)-J^*\sim 1/t$ \cite{nodecal-wright}. Whilst gradient descent is universally popular, alternative methods such as momentum and Nesterov's Accelerated Gradient (NAG) can result in significantly faster convergence to the optimum.

\subsection{Momentum}

The intuition behind momentum (MOM) is to continue updating the parameter along the previous update direction. This gives the algorithm (see for example \cite{momentum})
\begin{equation}
\begin{split}
v_{t+1} &= \mu_t v_t - \alpha_t \gradJ{\theta_t} \\
\theta_{t+1} &= \theta_t + v_{t+1} 
\end{split}
\label{eq:mom}
\end{equation}
where $0\leq\mu_t\leq 1$ is the momentum parameter. It is well known that GD can suffer from plateauing when the objective landscape has ridges (due to poor scaling of the objective, for instance) causing the optimization path to zig-zag. Momentum can alleviate this since persistent ascent directions accumulate in (2), whereas directions in which the gradient is quickly changing tend to cancel each other out.
The algorithm is also useful in the stochastic setting when only a sample of the gradient is available. By averaging the gradient over several minibatches/samples, the averaged gradient will better approximate the full batch gradient. In a different setting, when the objective function becomes flat, momentum is useful to maintain progress along directions of shallow gradient.  As far as we are aware, relatively little is known about the convergence properties of momentum. We show below, at least for a special quadratic objective, that momentum indeed converges.

\subsection{Nesterov's Accelerated Gradient}

Nesterov's Accelerated Gradient (NAG) \cite{nag} is given by 
\begin{equation}
\label{nag_org}
\begin{split}
y_{t+1} &= (1+\mu_t) \theta_{t} - \mu_t \theta_{t-1}\\
\theta_{t+1}  &= y_{t+1} - \alpha_t \gradJ{y_{t+1}}
\end{split}
\end{equation}
NAG has the interpretation that the previous two parameter values are smoothed and a gradient descent step is taken from this smoothed value. For Lipshitz convex functions (and a suitable schedule for $\mu_t$ and $\alpha_t$), NAG converges at rate $1/t^2$. Nesterov proved that this is the optimal possible rate for any method based on first order gradients\footnote{This is a `worst case' result. For example for quadratic functions, convergence is exponentially fast, leaving open the possibility that other algorithms may have superior convergence on `benign' problems. } \cite{nag}.  Nesterov proposed the schedule $\mu_t = 1-3/(5+t)$ and fixed $\alpha_t$, which we adopt in the experiments.\newl



Recently, \cite{ontheimportance} showed that by setting  $v_{t+1}=\theta_{t+1}-\theta_t$, \eqref{nag_org} can be rewritten as:
\begin{equation}
\begin{split}
v_{t+1} &= \mu_t v_t - \alpha_t \gradJ{\theta_t + \mu_t v_t} \\
\theta_{t+1} &= \theta_t + v_{t+1} 
\end{split}
\label{eq:nag}
\end{equation}
This formulation reveals the relation of NAG to the classical momentum algorithm \eqref{eq:mom} which uses $\gradJ{\theta_t}$ in place of $\gradJ{\theta_t + \mu_t v_t}$ in \eqref{eq:nag}. In both cases, NAG and MOM tend to continue updating the parameters along the previous update direction.\newl

In the machine learning community, NAG is largely viewed as somewhat mysterious and explained as performing a lookahead gradient evaluation and then performing a correction \cite{ontheimportance}.  The closely related momentum is often loosely motivated by analogy with a physical system \cite{momentum}. One contribution of our work, presented in \secref{sec:unified}, shows that these algorithms can be intuitively understood from the perspective of optimising the objective with respect to the update $v_t$ itself. 

\section{Regularised Update Descent\label{sec:unified}}

We consider a separable objective
\beq
\hat{J}(\theta_t,v_t)\equiv J(\theta_t)+\frac{\gamma}{2} v_t^2
\eeq
for which the $\theta$ that minimises $\hat{J}$ is clearly the same as the one that minimises $J$, with $v_t=0$ at the minimum.  
We propose\footnote{Previous authors have also considered optimising the update, for example \cite{masse2015speed}.} to update $\theta_t$ to $\theta_t+v_t$
to reduce $\hat{J}$. To do this we update $v_t$ to reduce\footnote{Note that the regulariser term $\gamma_tv_t^2/2$ is necessary. For the objective $J(\theta_t+v_t)$ alone, the update would be $v_{t+1}=v_t-\alpha_t J'(\theta_t+v_t)$. In this case, convergence for $v$ occurs when $ J'(\theta_t+v_t)=0$, for which $v_{t+1}=v_t$. Using the update $\theta_{t+1}=\theta_t+v_{t}$ would then result in the parameter $\theta$ never converging;  the parameter $\theta$ would pass though the minimum $J'(\theta)=0$ and continue beyond this, never to return.}  
\beq
\tilde{J}(\theta_t,v_t)\equiv \hat{J}(\theta_{t}+v_t,v_t)=J(\theta_t+v_t)+\frac{\gamma}{2} v_t^2
\eeq
We note that the optimum of $\tilde{J}$ occurs when
\begin{align}
\pdiff{\tilde{J}}{\theta_t}=0, \ocm \pdiff{\tilde{J}}{v_t}=0
\end{align}
These two conditions give
\begin{align}
J'(\theta_t+v_t)=0, \ocm J'(\theta_t+v_t)+\gamma_t v_t=0
\end{align}
which implies that at the optimum $v_t=0$ and therefore that $J'(\theta_t)=0$ when we have found the optimum of $\tilde{J}$. Hence, the $\theta_t$ that minimises $\tilde{J}$ also minimises $J$. \newl

Differentiating $\tilde{J}$ with respect to $v_t$ we obtain
\beq
\gradJ{\theta_t+v_t}+\gamma_t v_t
\eeq
We thus make a gradient descent update in the direction that lowers $\tilde{J}$:
\begin{align}
v_{t+1}&=v_t-\alpha_t\br{\gradJ{\theta_t+v_t}+ \gamma_t v_t}
\end{align}
We initially proposed to optimise $J(\theta)$ via the update $\theta_{t+1} = \theta_ t+ v_t$ by performing gradient descent on $\tilde J$ with respect to $v_t$. However, we have now improved $v_t$ to $v_{t+1}$. This suggests therefore that a superior update for $\theta_t$ is $\theta_{t+1}=\theta_t+v_{t+1}$.
The complete Regularised Update Descent (RUD) algorithm is given by (see also \algref{alg:rud})
\begin{equation}
\begin{split}
v_{t+1} &= \mu_t v_t - \alpha_t \gradJ{\theta_t + v_t} \\
\theta_{t+1} &= \theta_t + v_{t+1} 
\end{split}
\label{eq:rud}
\end{equation}
where $\mu_t\equiv 1-\alpha_t\gamma_t$. As we converge towards a minimum, the update $v_t$ will become small (since the gradient is small) and the regularisation term can be safely tuned down. This means that $\mu_t$ should be set so that it tends to 1 with increasing iterations. 
As we will show below one can view MOM and NAG as approximations to RUD based on a first order expansion (for MOM) and a more accurate second order expansion (for NAG).

\begin{algorithm}
  \caption{Regularised Update Descent for $T$ iterations
    \label{alg:rud}}
  \begin{algorithmic}[1]
    \Require{Initial guess $\theta_1$, learning rates $\alpha_t$ and increasing momentum schedule $0\leq \mu_t\leq 1$}
      \Let{$v_1$}{0} 
      \For{$t \gets 1 \textrm{ to } T-1$}
          \Let{$v_{t+1}$}{$\mu_t v_t-\alpha_t J'(\theta_t+v_t)$}
          \Let{$\theta_{t+1}$}{$\theta_t+v_{t+1}$}
      \EndFor
      \State \Return{$\theta_{T}$}
  \end{algorithmic}
\end{algorithm}


\subsection{Deriving MOM from RUD}

We consider an update $v_t$ at the current $\theta_t$. Assuming $v_t$ is small:
\beq
J(\theta_t+v_t)= J(\theta_t)+v_t\gradJ{\theta_t}+\bigo{v_t^2}
\eeq
Under this first order approximation, the RUD objective becomes
\beq
J(\theta_t)+v_t\gradJ{\theta_t}+\frac{\gamma_t}{2} v_t^2
\eeq
Differentiating wrt $v_t$ we get
\beq
\gradJ{\theta_t}+\gamma_t v_t
\eeq
We thus make an update in this direction:
\begin{align}
v_{t+1}&=v_t-\alpha_t\br{\gradJ{\theta_t}+\gamma_t v_t}\\
&=\mu_t v_t-\alpha_t\gradJ{\theta_t}
\end{align}
where $\mu_t$ should be close to 1.  We then make a parameter update
\beq
\theta_{t+1}=\theta_t+v_{t+1}
\eeq
which recovers the momentum algorithm. We can therefore view momentum as optimising, with respect to the update, a first order approximation of the RUD objective. 

\subsection{Deriving NAG from RUD}

Expanding $J(\theta_t+v_t)$ to the next order, we obtain
\beq
J(\theta_t+v_t)= J(\theta_t)+v_t\gradJ{\theta_t}+\frac{1}{2}v_t^2\gradgradJ{\theta_t}+\bigo{v_t^3}
\eeq
Since $v_t$ is not infinitesimally small, we cannot `trust' the higher order terms as we move away from $v_t=0$; as we move further from $\theta_t$ we are trying to approximate the function based on curvature information at $\theta_t$, rather than the current point $v_t+\theta_t$. This is analogous to the idea of trust regions in Quasi-Newton approaches which limit the extent to which the Taylor expansion is trusted away from the origin \cite{nodecal-wright}. To encode this lack of trust, we reduce the second order term by a factor $\mu_t<1$ and add another term to encourage $v_t$ to be small. This gives the modified approximate RUD objective
\beq
J(\theta)+v_t\gradJ{\theta_t}+\frac{\mu_t}{2}v_t^2\gradgradJ{\theta}+\frac{\gamma_t}{2} v_t^2
\eeq
Differentiating with respect to $v_t$ we get
\beq
\gradJ{\theta_t}+\mu_t v_t\gradgradJ{\theta_t}+\gamma_t v_t=\gradJ{\theta_t+\mu_t v_t} + \gamma_t v_t + \bigo{v^2}
\eeq
We then update $v_t$ to reduce this approximate RUD objective:
\begin{align}
v_{t+1}&=v_t-\alpha_t\br{\gradJ{\theta_t+\mu_t v_t} + \gamma_t v_t}\\
&=(1-\alpha_t\gamma_t)v_t-\alpha_t\gradJ{\theta_t+\mu_t v_t}
\end{align}
We are free to choose $\alpha_t$, and $\gamma_t$ which should both be small. Ideally $\mu_t$ should be close to 1. Hence, it is reasonable to set $1-\alpha_t\gamma_t=\mu_t$ and choose $\mu_t$ to be close to 1. This setting  recovers the NAG algorithm:
\begin{align}
v_{t+1}&=\mu_t v_t-\alpha_t\gradJ{\theta_t+\mu_t v_t}\\
\theta_{t+1}&=\theta_t+v_{t+1}
\end{align}
and explains why we want $\mu_t$ to tend to 1 as we converge, since as we zoom in to the minimum, we can trust more a quadratic approximation to the objective.  An alternative interpretation of NAG (as a two stage optimisation process) and its relation to RUD is outlined in \appref{sec:alg:nag}. \newl

From the perspective that NAG and MOM are approximations to RUD, NAG is preferable to MOM since it is based on a more accurate expansion. In terms of RUD versus NAG, the difference between NAG and RUD is the use of $\mu_t$ in the argument of $\gradJ{\theta_t+\mu_tv_t}$ in NAG, whereas we use  $\gradJ{\theta_t+v_t}$ in RUD. This means that RUD `looks further forward' than NAG (since $\mu_t<1$) in a manner more consistent with the eventual parameter update $\theta_t+v_{t+1}$. This tentatively explains why RUD can outperform NAG.

\section{Comparison on a Quadratic function\label{sec:analysis}}

An interesting question is whether and under what conditions RUD may converge more quickly than NAG for convex Lipshitz functions. To date we have not been able to fully analyse this. In lieu of a more complete understanding we consider the simple quadratic objective\footnote{For the simple quadratic objective, the convergence is exponentially fast in terms of the number of iterations. This is clearly a very special case compared to the more general convex Lipshitz scenario. Nevertheless, the analysis gives some insight that some improvement over NAG might be possible.}
\beq
J(\theta)= \frac{1}{2}\theta^2
\eeq
For this simple function the gradient is given by $\theta$ and, for fixed $\alpha_t$, $\mu_t$, we are able to fully compute the update trajectories for NAG and RUD and MOM.

\subsection{NAG}

For NAG, the algorithm is given by
\begin{align}
v_{t+1} &= \mu v_t -\alpha(\theta_t+\mu v_t)\label{eq:nag:update:one}\\
\theta_{t+1}&=\theta_t+v_{t+1}
\label{eq:nag:update}
\end{align}
Assuming $v_1=0$, and a given value for $\theta_1$, this gives $\theta_2=(1-\alpha)\theta_1$. Similarly, for both MOM and NAG, $\theta_2$ is given by the same value.\newl

\begin{figure}[t]
\begin{center}
\subfigure[]{\includegraphics[height=0.25\tw]{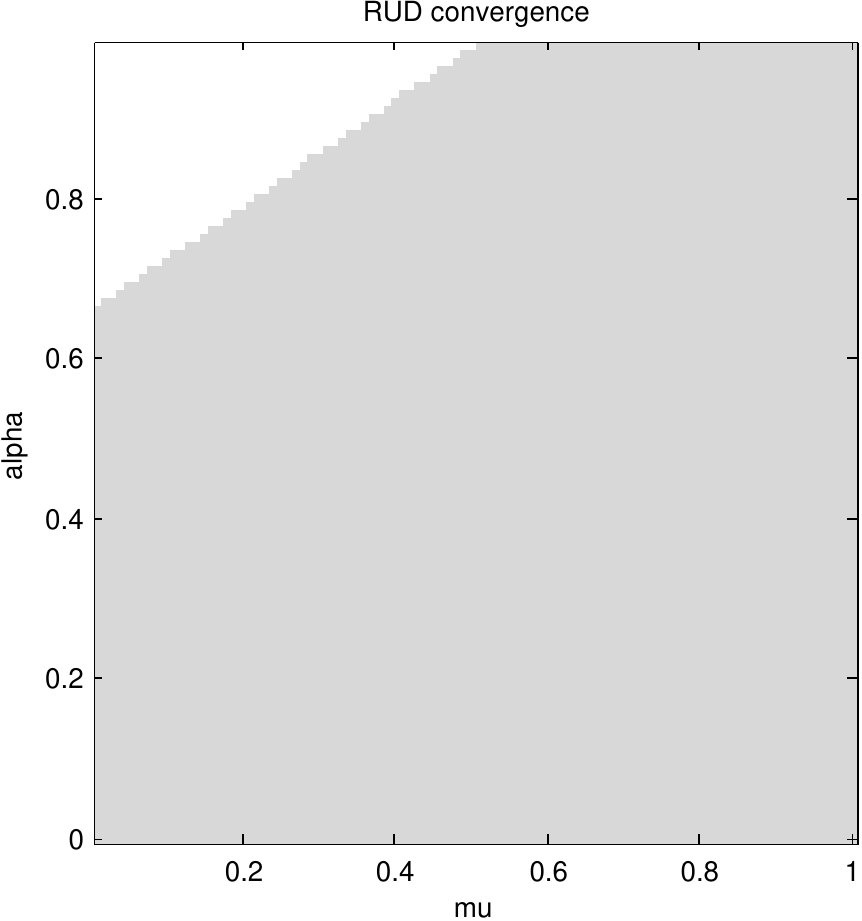}}
\subfigure[]{\includegraphics[height=0.25\tw]{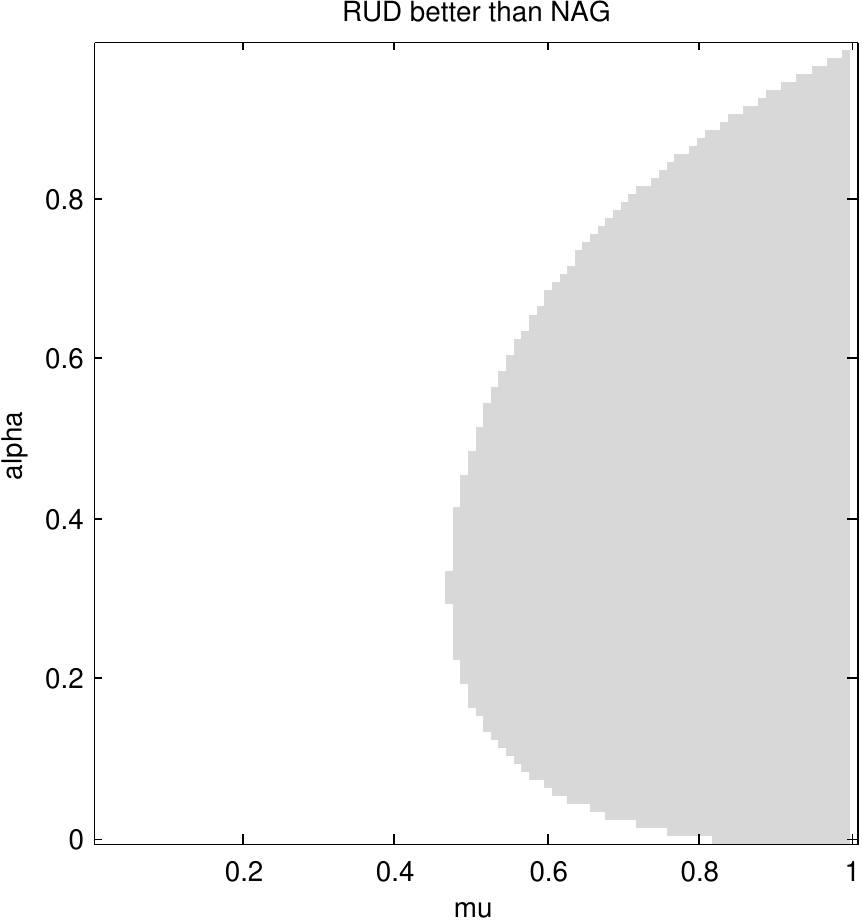}}
\subfigure[]{\includegraphics[height=0.25\tw]{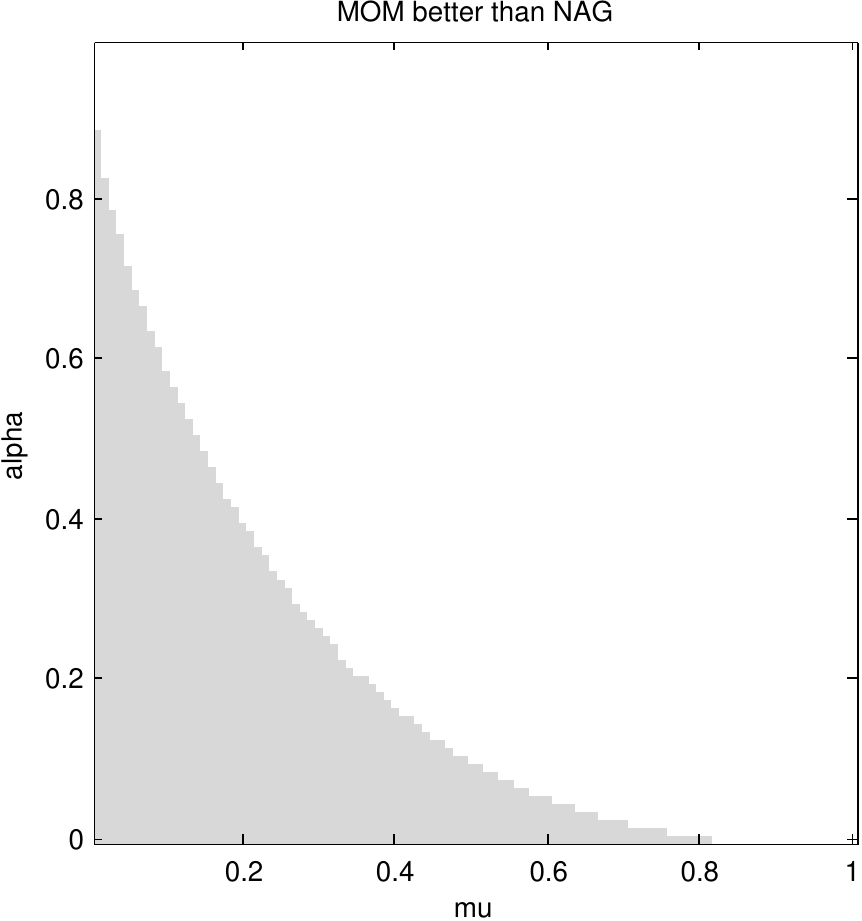}}
\subfigure[]{\includegraphics[height=0.25\tw]{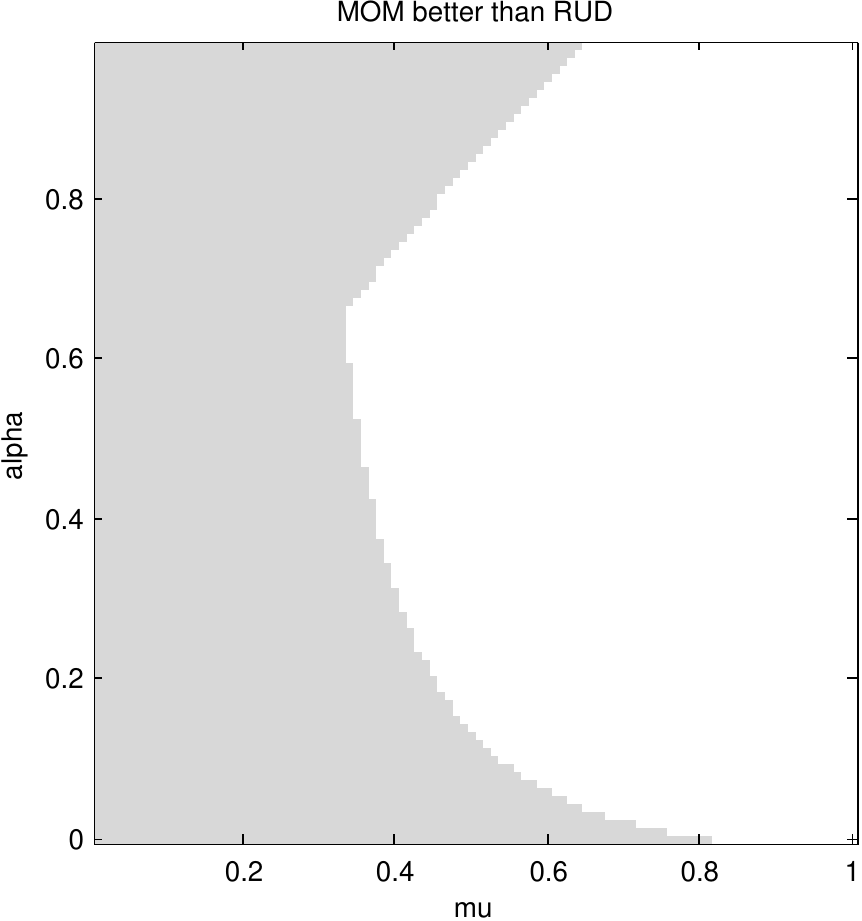}}
\end{center}
\caption{(a) Shaded is the parameter region $(\mu,\alpha)$ for which RUD converges for the simple quadratic function $f(\theta)=0.5\theta^2$. (b) Shaded is the parameter region $(\mu,\alpha)$ for which RUD converges more quickly than NAG. (c) Shaded is the region in which MOM converges more quickly than NAG. (d) Shaded is the region in which MOM converges more quickly than RUD.\label{fig:quad:conv}}
\end{figure}

We can write equations (\ref{eq:nag:update:one},\ref{eq:nag:update}) as a single second order difference equation
\beq
\theta_{t+1}+b\theta_t+c\theta_{t-1} = 0
\eeq
where
\begin{align}
b &\equiv -1-\mu+\alpha+\alpha\mu\\
c&\equiv \mu-\alpha\mu 
\end{align}
For the scalar case $\dim(\theta)=1$, assuming a solution of the form $\theta_t=Aw^t$ gives
\beq
w = \frac{-b\pm \sqrt{b^2-4c}}{2}
\eeq
which defines two values $w_+$ and $w_-$, so that the general solution is given by
\beq
\theta_t = Aw_+^t + Bw_-^t
\eeq
where $A$ and $B$ are determined by the linear equations
\begin{align}
\theta_1&=Aw_++Bw_-\\
\theta_2&=Aw_+^2+Bw_-^2
\end{align}
A sufficient condition for NAG to converge is that $|w_+|<1$ and $|w_-|<1$ which is equivalent to the conditions $|b|<1+c$, $c<1$ \cite{SydsaeterHammondBook}.  For any learning rate $0<\alpha<1$ and momentum $0<\mu<1$, it is straightforward to show that these conditions hold and thus that NAG converges to the minimum $\theta_*=0$.

\begin{figure}[t]
\begin{center}
\subfigure[]{\includegraphics[height=0.45\tw]{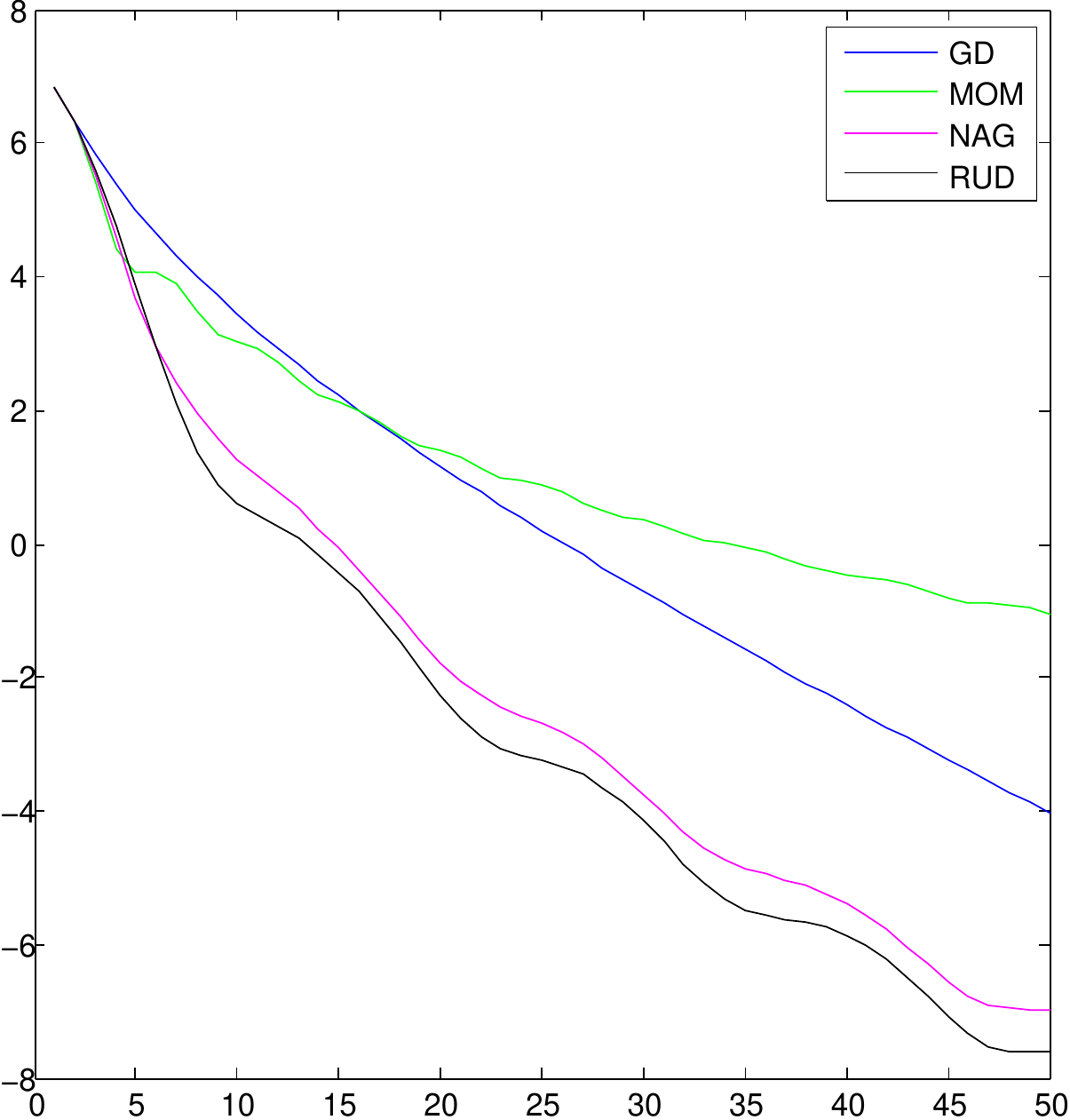}}\hcm
\subfigure[]{\includegraphics[height=0.45\tw]{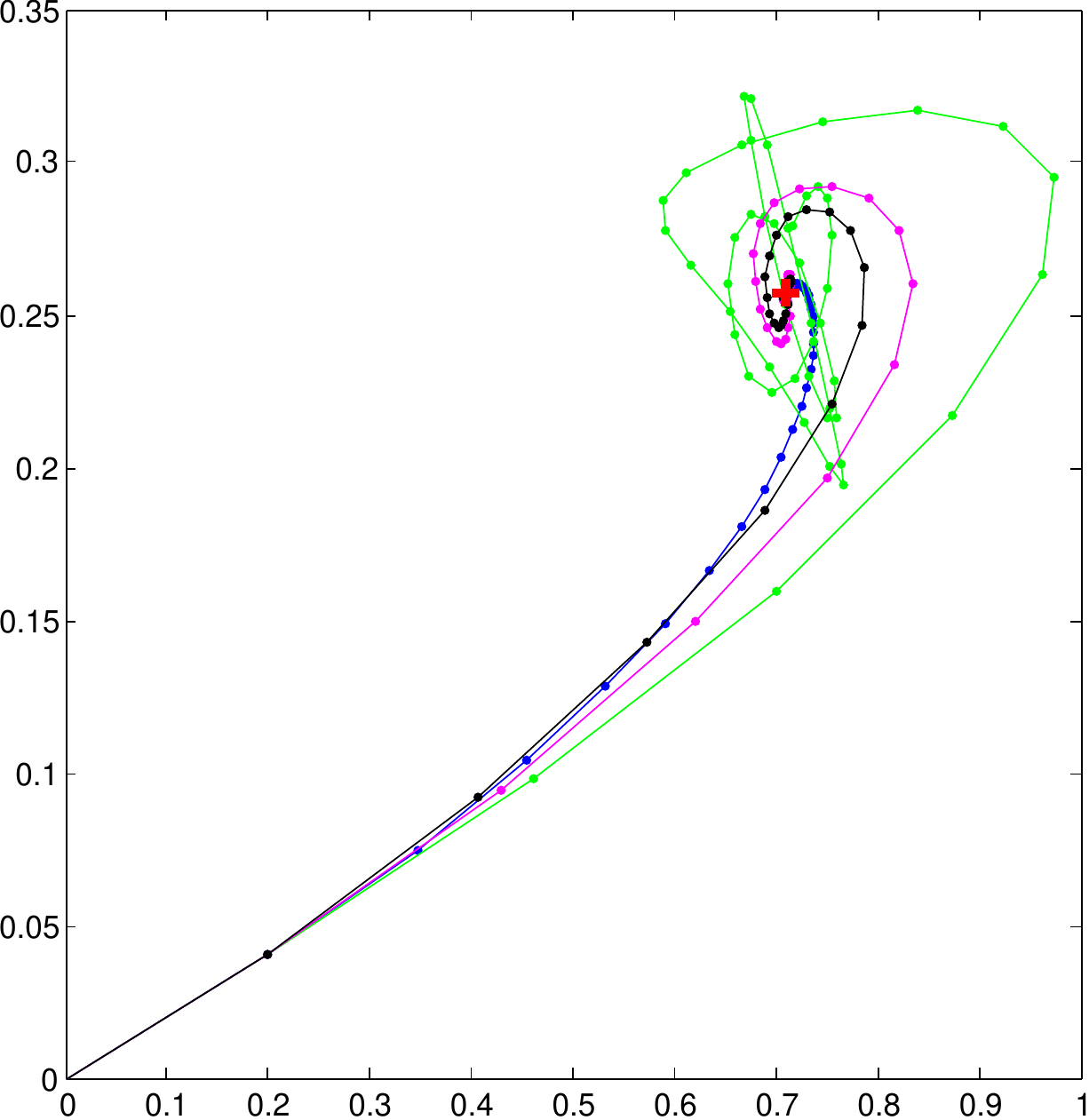}}
\end{center}
\caption{Optimising a 1000 dimensional quadratic function $J(\theta)$ using different algorithms, all with the same learning rate $\alpha_t$ and $\mu_t$ schedule. (a) The log objective $\log J(\theta_t)$  for Gradient Descent, Momentum, Nesterov's Accelerated Gradient and Regularised Update Descent. (b) Trajectories of the different algorithms plotted for the first two components $(\theta_{1t},\theta_{2t})$. The behaviour demonstrated is typical in that momentum tends to more significantly overshoot the minimum than RUD or NAG, with RUD typically outperforming NAG.\label{fig:quad}}
\end{figure}

\subsection{MOM}

The above analysis carries over directly to the MOM algorithm, with the only change being 
\begin{align}
b &\equiv -1-\mu+\alpha\\\
c&\equiv \mu 
\end{align}
It is straightforward to show that for any learning rate $0<\alpha<1$ and momentum $0<\mu<1$, the corresponding conditions $|w_+|<1$ and $|w_-|<1$ are always satisfied. Therefore the MOM algorithm (at least for this problem) always converges. For MOM to have better asymptotic convergence rate than NAG, we need $\max(|w^{MOM}_+|,|w^{MOM}_-|) < \max(|w^{NAG}_+|,|w^{NAG}_-|)$. From \figref{fig:quad:conv} we see that MOM only outperforms NAG (and RUD) when the momentum is small. This is essentially the uninteresting regime since, in practice, we will typically use a value of momentum that is close to 1. For this simple quadratic case, for practical purposes, MOM therefore performs worse than RUD or NAG. 

\subsection{RUD}

For the RUD algorithm the corresponding solutions are given by setting
\begin{align}
b &\equiv -1-\mu+2\alpha\\\
c&\equiv \mu-\alpha 
\end{align}
RUD has more complex convergence behaviour than NAG or MOM. The conditions  $|w_+|<1$ and $|w_-|<1$ are satisfied only within the region as shown in \figrefm{fig:quad:conv}{a}, which is determined by
\beq
1+\mu > \frac{3}{2}\alpha
\eeq
The main requirement is that the learning rate should not be too high, at least for values of momentum $\mu$ less than 0.5. Unlike NAG and MOM, RUD has therefore the possibility to diverge.\newl

In \figrefm{fig:quad:conv}{b} we show the region for which the asymptotic convergence of RUD is faster than NAG.  The main requirement is that the momentum needs to be high (say above 0.8) and is otherwise largely independent of the learning rate (provided $\alpha<1$). 

\section{Experiments\label{sec:experiments}}

\subsection{A toy high dimensional quadratic function}
In \figref{fig:quad} we show the progress for different algorithms using the same learning rate $\alpha_t=0.2$ and $\mu_t=1-3/(5+t)$ for a toy 1000 dimension quadratic function $\frac{1}{2}\theta\trans A \theta-\theta\trans b$ for randomly chosen $A$ and $b$. This simple experiment shows that the theoretical property derived in \secref{sec:analysis} that RUD can outperform NAG and MOM carries over to the more general quadratic setting. Indeed, in our experience, the improved convergence of RUD over NAG for the quadratic objective function is typical behaviour.

\subsection{Deep Learning: MNIST}

\begin{figure}[t]
\bmp{0.5\tw}{\includegraphics[height=0.85\tw]{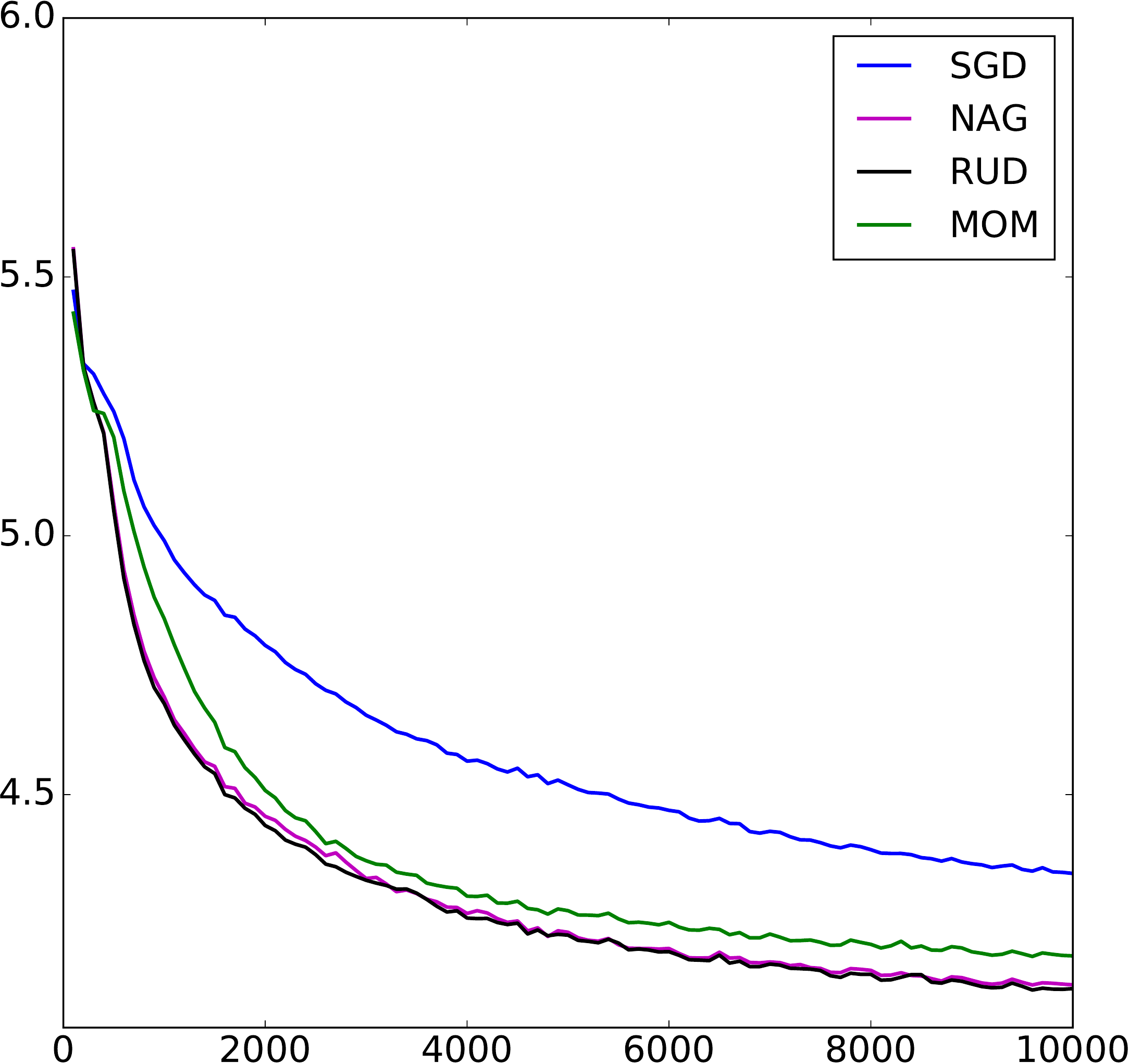}}\emp
\bmp{0.5\tw}\caption{The negative log loss for the classical MNIST $784-1000-500-25-30$ autoencoder network \cite{hinton_arch} trained using minibatches contains 200 examples.  Similar to the small quadratic objective experiments, we see that on this much larger problem, as expected, NAG and RUD perform very similarly (with RUD slightly outperforming NAG). All methods used the same learning rate and momentum parameter $\mu_t$ schedule.  \label{fig:mnist}}\emp
\end{figure}

Whilst RUD has interesting convergence for quadratic functions, in practice of course it is important to see how it behaves in the case of more general non-convex functions. In \figref{fig:mnist} we look at a classical deep learning problem of training an  $784-1000-500-250-30$ autoencoder for handwritten digit reconstruction \cite{hinton_arch}.   The dataset consists of black and white images of size 28x28 and we used 50000 training images, with the images scaled to lie in the 0 to 1 range.  The target is for the network to learn to reduce the dimensionality of the input to a 30 dimensional vector and then to reconstruct the input. The nonlinearity at each layer is the hyperbolic tangent\footnote{We tried also rectifier linear units and leaky rectifier linear units, but they did not affect the relative performance of any of the algorithms.} and for the last layer we used the binary cross entropy loss. \newl

Since NAG and RUD are closely related, we use the same schedule $\mu_t = 1-3/(5+t)$ for both algorithms. All remaining hyperparameters for each method (learning rates) were set optimally based on a grid search over a set of reasonable parameters for each algorithm.   For this problem, there is little difference between NAG and RUD, with RUD slightly outperforming NAG.  

\section{Conclusion}

We described a general approach to first order optimisation based on optimising the objective with respect to the updates. This gives a simple optimisation algorithm which we termed Regularised Update Descent; we showed that his algorithm can converge more quickly than Nesterov's Accelerated Gradient.  
In addition to being  a potentially useful optimisation algorithm in its own right, the main contribution of this work is to show that the Nesterov and momentum algorithms can be viewed as approximations to the Regularised Update Descent algorithm.  

%
%


\bibliographystyle{plain}
\bibliography{references}

\appendix
\section{Alternative NAG derivation\label{sec:alg:nag}}

For the objective
\begin{equation}
\label{regularized_form}
\tilde{J}(\theta_t,v_t) = J(\theta_t+v_t) + \frac{1}{2}\gamma_t v_t^2
\end{equation}
we consider a two stage process of optimizing $\tilde{J}$. The algorithm proceeds as follows: given $\theta_t$ and $v_t$ we first perform a descent step only on the regularizer, followed by a descent step on the `lookahead' $J(\theta_t + v)$. After this we perform the usual step on $\theta_t$ based on the final updated $v$. The procedure is summarized below:
\begin{equation}
\begin{split}
\tilde{v}_{t+1} &= v_t - \alpha_t \gamma_t v_t = (1 - \alpha_t \gamma_t) v_t \\
g_t &= J'(\theta_t + \tilde{v}_{t+1}) \\
v_{t+1} &= \widetilde{v}_{t+1} - \alpha_t g_t = (1 - \alpha_t \gamma_t) v_t - \alpha_t g_t \\
\theta_{t+1} &= \theta_t + v_{t+1} 
\end{split}
\end{equation}
Setting  $\mu_t=1- \alpha_t\gamma_t$ recovers the NAG formulation as in \cite{ontheimportance}. RUD therefore differs from NAG in that it does not perform the initial descent step on the regulariser term so that for RUD $\tilde{v}_{t+1}=v_t$. 


\end{document}